\documentclass{article}
\usepackage[preprint,nonatbib]{neurips_2019}

\usepackage[utf8]{inputenc} % allow utf-8 input
\usepackage[T1]{fontenc}    % use 8-bit T1 fonts
\usepackage{url}            % simple URL typesetting
\usepackage{booktabs}       % professional-quality tables
\usepackage{amsfonts}       % blackboard math symbols
\usepackage{nicefrac}       % compact symbols for 1/2, etc.
\usepackage{microtype}      % microtypography
\usepackage{graphicx}
\usepackage{subfigure}
\usepackage{amsmath}
\usepackage{multirow}
\usepackage{array}
\usepackage{algorithm}
\usepackage{algpseudocode}
\title{Automatic and Simultaneous Adjustment of Learning Rate and Momentum
for Stochastic Gradient Descent}

\author{%
  Tomer Lancewicki\\
  eBay\\
  \texttt{tlancewicki@ebay.com} \\
  \And
  Selcuk Kopru\\
  eBay\\
  \texttt{skopru@ebay.com}\\
}

\begin{document}

\maketitle

\begin{abstract}
Stochastic Gradient Descent (SGD) methods are prominent for training machine learning and deep learning models. The performance of these techniques depends on their hyperparameter tuning over time and varies for different models and problems. Manual adjustment of hyperparameters is very costly and time-consuming, and even if done correctly, it lacks theoretical justification which inevitably leads to ``rule of thumb'' settings. In this paper, we propose a generic approach that utilizes the statistics of an unbiased gradient estimator to automatically and simultaneously adjust two paramount hyperparameters: the learning rate and momentum. We deploy the proposed general technique for various SGD methods to train Convolutional Neural Networks (CNN's). The results match the performance of the best settings obtained through an exhaustive search and therefore, removes the need for a tedious manual tuning. 
\end{abstract}

\section{Introduction}
\label{sec:introduction}
Machine learning has intimate ties to optimization, considering that
many learning problems are formulated as the minimization of a loss
function that depends on a training set. An optimization problem that
frequently appears in machine learning is the minimization of the
average of loss functions over a finite training set, i.e.,
\begin{equation}
\bar{F}\left(\mathbf{w}\right)=\frac{1}{M}\sum_{i=1}^{M}f\left(\mathbf{w};\bar{\mathbf{x}}_{i}\right),\label{eq:loss_F}
\end{equation}
where $\bar{\mathbf{x}}_{i}\in\mathbb{R}^{d}$ is the $i$-th observation in the training set $\left\{ \bar{\mathbf{x}}_{i}\right\} _{i=1}^{M}$ of size $M$, the function $f\left(\mathbf{w};\bar{\mathbf{x}}_{i}\right):\mathbb{R}^{d}\rightarrow\mathbb{R}$ is the loss corresponding to $\bar{\mathbf{x}}_{i}$, and $\mathbf{w}\in\mathbb{R}^{p}$
is the weight vector. Starting with an initial guess for the weight
vector $\mathbf{w}$, \emph{stochastic gradient descent} (SGD) methods
\cite{spall2005introduction} attempt to minimize the loss function
$\bar{F}\left(\mathbf{w}\right)$ \eqref{eq:loss_F} by iteratively
updating the values of $\mathbf{w}$. Each iteration utilizes a sample
$\left\{ \mathbf{x}_{i}\right\} _{i=1}^{N}$ of size $N$, commonly
called a ``mini-batch'', which is taken randomly from the training
set $\left\{ \bar{\mathbf{x}}_{i}\right\} _{i=1}^{M}$. The update
from $\mathbf{w}_{t}$ to $\mathbf{w}_{t+1}$ at the $t$-th iteration
relies on a gradient estimator, which in turn depends on the current
mini-batch. The typical unbiased gradient estimator of the unknown
true gradient is defined by
\begin{equation}
\mathbf{g}_{t}=\frac{1}{N}\sum_{i=1}^{N}\frac{\partial f\left(\mathbf{w};\mathbf{x}_{i}\right)}{\partial\mathbf{w}}\mid_{\mathbf{w}=\mathbf{w}_{t}}=\frac{1}{N}\sum_{i=1}^{N}\mathbf{g}_{t}^{\left(i\right)},\label{eq:gt}
\end{equation}
where $\mathbf{g}_{t}^{\left(i\right)}$ is the gradient produced
by the $i$-th observation within the current mini-batch of size $N$.
The gradient estimator $\mathbf{g}_{t}$ \eqref{eq:gt} entails variance,
since it depends on a random set of observations. If the variance
of $\mathbf{g}_{t}$ \eqref{eq:gt} is large, the SGD method may have
difficulty converging and perform poorly. Indeed, the variance may
be reduced by increasing the mini-batch size $N$. However, this increases
the computational cost of each iteration. Some recent methods in the
literature that attempt to reduce the variance of the gradient estimator
include \cite{NIPS2013_5034,johnson2013accelerating,reddi2015variance,icassp1,icassp5},
to mention a few. While these methods provide unbiased gradient estimators,
they are not necessarily optimal in the sense of \emph{mean-squared
error} (MSE) which allows reducing the variance with the cost of bias.
Momentum-based methods (see \cite{sutskever2013importance} and other
references within) trade-off between variance and bias by constructing
the gradient estimator as a combination of the current unbiased gradient
estimator and previous gradient estimators. Other state-of-the-art
methods use biased estimators by scaling the gradient with square
roots of exponential moving averages of past squared gradients \cite{j.2018on}.
These methods include, for example, AdaGrad \cite{duchi2011adaptive},
Adam \cite{kingma2014adam}, AdaDelta \cite{zeiler2012adadelta},
NAdam \cite{dozat2016incorporating}, etc. The main drawback for these
methods is their reliance on one or more hyperparameters, i.e., parameters
which must be tuned in advance to obtain adequate performance. Unfortunately,
manual hyperparameter tuning is very costly, as every hyperparameter
configuration is typically tested over many iterations. Previous attempts to automatically tune the learning rate alone were proposed in \cite{schaul2013no,tan2016barzilai} and examined for simple architectures, such as \emph{logistic regression} and \emph{fully-connected neural networks} (FCNN's). The technique presented in \cite{schaul2013no},
for example, proposes an automatic adjustment of the learning rate,
limited by the assumption of a diagonal Hessian matrix, and disregarding
the off-diagonal elements of the observations' covariance matrix.
The approach proposed in \cite{tan2016barzilai} set the learning
rate by utilizing the Barzilai-Borwein method \cite{barzilai1988two}
which in turn relies on an approximation of the Hessian, ignoring
gradient estimators as suggested in \cite{schaul2013no}. To the best
of our knowledge, there is no method to adjust the momentum hyperparameter
automatically; neither in solitary nor simultaneously with the learning
rate. 

\begin{figure}
  \centering
  \includegraphics[width=0.8\textwidth]{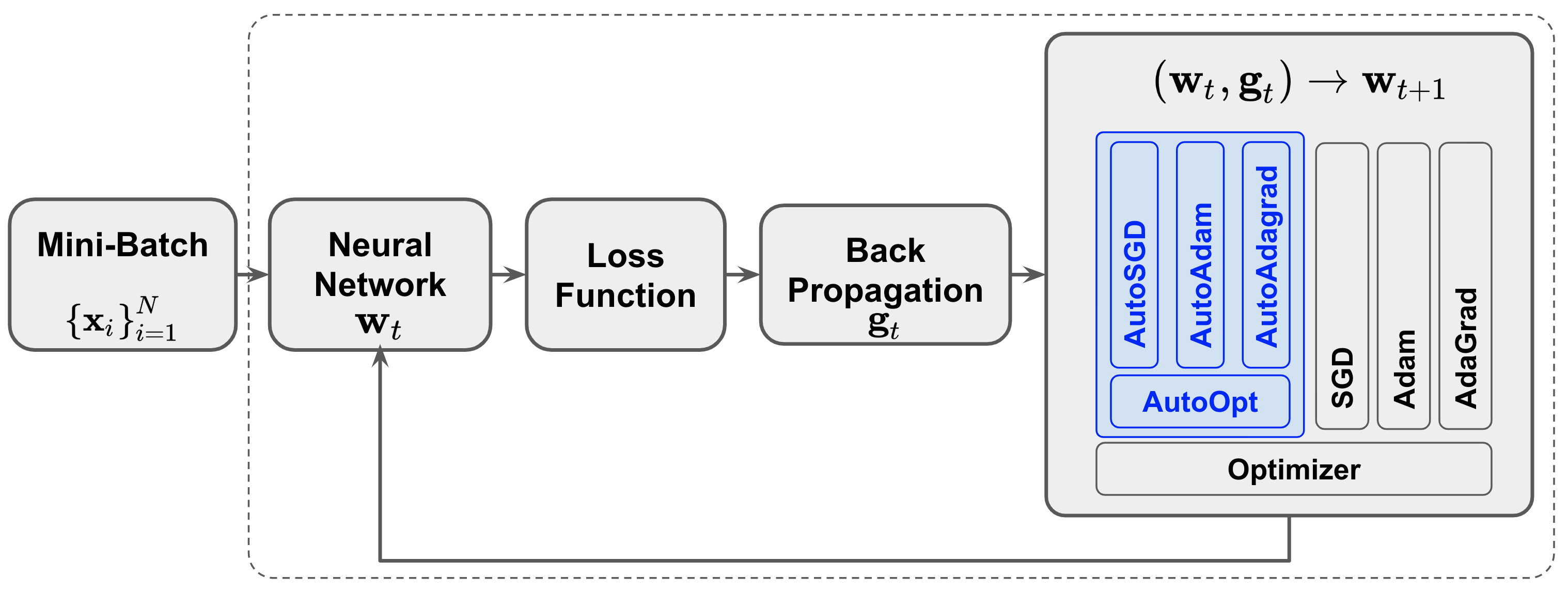}
  \caption{The proposed method, dubbed as AutoOpt, provides a general approach to an automatic and simultaneous adjustment of the learning rate and momentum hyperparameters. We deploy and examine the generic technique for the SGD, Adam, and AdaGrad optimizers, for the sake of training CNN based classifiers. %The AutoOpt module is available in \cite{AutoOpt}.
  }
  \label{fig:Architecture}
\end{figure}

In this paper, we present a novel and generic method to automatically and simultaneously adjust the learning rate and the momentum hyperparameters, to minimize (or maximally decrease) the expected loss after the next update. The general method, dubbed as AutoOpt, is deployed for three popular optimizers: SGD, Adam and AdaGrad, schematically described in Figure \ref{fig:Architecture}. The technique is practical for modern deep learning architectures and is successfully examined for \emph{convolutional neural networks} (CNN's) \cite{NIPS2012_4824}. The rest of the paper is organized as follows. In Section
2, we provide background and motivation for the proposed method. In
Section 3, we derive the general formulation and the theoretical properties
of the optimal learning rate and momentum. The optimal values depend
on the unknown true gradient, thus unattainable and annunciate as
the ``oracle'' solution. Nevertheless, we show in Section 4 that
the oracle solution can be estimated, hence makes it feasible for a
practical use. We deploy the proposed technique in SGD, Adam \cite{kingma2014adam}, and AdaGrad \cite{duchi2011adaptive} for the sake of training 
CNN based classifiers. Our experimental results which appear in Section 5, show that the method automatically achieves the lowest or comparable classification errors, obtained through a tedious systematic search of the learning rate and
momentum. %The source code for the proposed method is provided in \cite{AutoOpt}. 
We conclude the paper with a discussion on some future work. \textbf{ Notations:} We depict vectors
in lowercase boldface letters and matrices in uppercase boldface.
The transpose operator and the diagonal operator are denoted by  $\left(\cdot\right)^{T}$ and $diag\left(\cdot\right)$, respectively. The column vector of $p$ ones is denoted by $\mathbf{1}_{p}=\left[1,1,\ldots,1\right]^{T}$
and the expectation operator is denoted by $E\left\{ \cdot\right\} $.

\section{Motivation}
\label{sec:motivation}

We outset our discussion from the theoretical (and ideal) scenario of an unlimited training set. Suppose that the observations within the training set $\left\{ \bar{\mathbf{x}}_{i}\right\} _{i=1}^{M}$ are \emph{independent identically distributed} (i.i.d), drawn from a probability density function $\mathcal{P}\left(\mathbf{x}\right)$.
In the ideal case of an unlimited amount of training examples, the loss function $\bar{F}\left(\mathbf{w}\right)$ \eqref{eq:loss_F} approaches
the real unknown loss function, defined as
\begin{equation}
J\left(\mathbf{w}\right)=\lim_{M\rightarrow\infty}\frac{1}{M}\sum_{i=1}^{M}f\left(\mathbf{w};\bar{\mathbf{x}}_{i}\right)=\int f\left(\mathbf{w};\mathbf{x}\right)\mathcal{P}\left(\mathbf{x}\right)d\mathbf{x}.
\label{eq:J_func}
\end{equation}
The loss function $J\left(\mathbf{w}\right)$ \eqref{eq:J_func} is deterministic, and assumed to be continuously
differentiable with respect to the weight vector $\mathbf{w}$. Starting
with an initial guess $\mathbf{w}_{0}$, we would like to generate
a sequence of weights $\mathbf{w}_{t},t=1,\ldots,T$ such that the
loss function $J\left(\mathbf{w}\right)$ \eqref{eq:J_func} is reduced at each iteration
of the algorithm, i.e.,
\begin{equation}
J\left(\mathbf{w}_{t+1}\right)<J\left(\mathbf{w}_{t}\right).
\end{equation}
The loss function $J\left(\mathbf{w}\right)$ \eqref{eq:J_func} can be approximated
by a second-order (i.e., quadratic) Taylor series expansion around
$\mathbf{w}_{t}$ , i.e.,
\begin{equation}
\begin{array}{c}
J\left(\mathbf{w}\right)\approx\hat{J}\left(\mathbf{w}\right)=J\left(\mathbf{w}_{t}\right)+\left(\mathbf{w}-\mathbf{w}_{t}\right)^{T}\mathbf{\bar{g}}_{t}\\
+\frac{1}{2}\left(\mathbf{w}-\mathbf{w}_{t}\right)^{T}\mathbf{\bar{H}}_{t}\left(\mathbf{w}-\mathbf{w}_{t}\right),
\end{array}\label{eq:Taylor-3}
\end{equation}
where
\begin{equation}
\mathbf{\bar{g}}_{t}=\frac{\partial J\left(\mathbf{w}\right)}{\partial\mathbf{w}}\mid_{\mathbf{w}=\mathbf{w}_{t}}\label{eq:real_gradient}
\end{equation}
and
\begin{equation}
\mathbf{\bar{H}}_{t}=\frac{\partial^{2}J\left(\mathbf{w}\right)}{\partial\mathbf{w}^{2}}\mid_{\mathbf{w}=\mathbf{w}_{t}}\label{eq:real_hessian}
\end{equation}
are the gradient vector and the Hessian matrix of the loss function
$J\left(\mathbf{w}\right)$ \eqref{eq:J_func}, evaluated at $\mathbf{w}_{t}$. By deriving
$\hat{J}\left(\mathbf{w}\right)$ \eqref{eq:Taylor-3} with respect
to $\mathbf{w}$ and setting the result to zero, we find that the
next weight vector $\mathbf{w}_{t+1}$ which minimizes $\hat{J}\left(\mathbf{w}\right)$
\eqref{eq:Taylor-3} is given by
\begin{equation}
\mathbf{w}_{t+1}=\mathbf{w}_{t}-\mathbf{\bar{H}}_{t}^{-1}\mathbf{\bar{g}}_{t}.\label{eq:Taylor-3-1}
\end{equation}
The iterative equation \eqref{eq:Taylor-3-1} is also known as the
Newton-Raphson method \cite{z1}. %\cite[Ch. 4.3.3]{z1}. 
 In practice, at time
$t$, only a finite sample (the current mini-batch) of size $N$ is
available. As a result, neither the gradient vector $\mathbf{\bar{g}}_{t}$
\eqref{eq:real_gradient}, nor the Hessian matrix $\mathbf{\bar{H}}_{t}$
\eqref{eq:real_hessian} (and its inverse) required in \eqref{eq:Taylor-3-1},
are known. To practically apply the update rule \eqref{eq:Taylor-3-1} for $\mathbf{w}_{t+1}$,
both quantities $\mathbf{\bar{g}}_{t}$ and $\mathbf{\bar{H}}_{t}$,
must be replaced by their estimators, denoted as $\mathbf{\hat{g}}_{t}$
and $\mathbf{\hat{H}}_{t}$, respectively. The use of the estimators
$\mathbf{\hat{g}}_{t}$ and $\mathbf{\hat{H}}_{t}$ (instead of $\mathbf{\bar{g}}_{t}$
and $\mathbf{\bar{H}}_{t}$) leads to the general update rule of SGD
methods given by
\begin{equation}
\mathbf{w}_{t+1}=\mathbf{w}_{t}-\hat{\mathbf{H}}_{t}^{-1}\mathbf{\hat{g}}_{t}.\label{eq:w_update-1}
\end{equation}
We confine our discussion regarding the gradient estimator
$\mathbf{\hat{g}}_{t}$ in \eqref{eq:w_update-1}, to the frequently used model
\begin{equation}
\mathbf{\hat{g}}_{t}=\alpha\left(\left(1-\beta\right)\mathbf{g}_{t}+\beta\mathbf{\hat{g}}_{t-1}\right),\label{eq:g_est}
\end{equation}
where $\mathbf{g}_{t}$ \eqref{eq:gt} is the unbiased estimator of
the unknown true gradient $\mathbf{\bar{g}}_{t}$ \eqref{eq:real_gradient} (i.e.,
$E\left\{ \mathbf{g}_{t}\right\} =\mathbf{\bar{g}}_{t}$), $\beta$
is the momentum parameter which is a scalar between 0 and 1, and $\alpha$
is the learning rate, a positive scalar which ensures that the update
rule \eqref{eq:w_update-1} do not produce a weight vector $\mathbf{w}_{t+1}$
with an implausible large norm \cite{bordes2009sgd}. The inverse of the estimated Hessian matrix $\hat{\mathbf{H}}_{t}^{-1}$
in \eqref{eq:w_update-1} is not easy to compute. There are various
methods to estimate the inverse of the Hessian matrix, such as \emph{Broyden-Fletcher-Goldfarb-Shanno}
(BFGS) quasi-Newton based methods \cite{mokhtari2014res,icassp2}.
However, if computational simplicity is of paramount importance, then
it is common to assume that $\hat{\mathbf{H}}_{t}^{-1}$ is equal
to the identity matrix $\mathbf{I}$. In this paper we refer to that case as the classic SGD. The Adam optimizer \cite{kingma2014adam}, a popular SGD algorithm which is vastly used these days, assumes a diagonal Hessian matrix estimator of the form 
\begin{equation}
\hat{\mathbf{H}}_{t}=\left(1-\beta^{t}\right)diag\left(\sqrt{\frac{\left(1-\beta_{2}\right)\mathbf{g}_{t}^{2}+\beta_{2}\mathbf{\hat{g}}_{t-1}^{2}}{1-\beta_{2}^{t}}}+\epsilon\mathbf{1}_{p}\right).\label{eq:Hessian_Adam}
\end{equation}
The AdaGrad method \cite{duchi2011adaptive}, which corresponds to a version of Adam, utilizes the gradient estimator $\mathbf{\hat{g}}_{t}$  \eqref{eq:g_est} with momentum $\beta=0$, and a diagonal Hessian matrix estimator of the form
\begin{equation}
\hat{\mathbf{H}}_{t}=diag\left(\sqrt{\sum_{i=1}^{t}\mathbf{g}_{i}^{2}}+\epsilon\mathbf{1}_{p}\right).\label{eq:Hessian_AdaGrad}
\end{equation} 
These methods propose different gradient and Hessian estimators, to be plugged into the update rule \eqref{eq:w_update-1}, and are summarized in Table 1. 
\begin{table}[t]
\caption{SGD methods that follow the general update rule \eqref{eq:w_update-1} along with their gradient and Hessian estimators. These optimizers share the same gradient estimator model $\mathbf{\hat{g}}_{t}$ \eqref{eq:g_est}, with or without momentum, while utilizing different Hessian estimators.}
\label{sample-table}
\vskip 0.15in
\begin{center}
\begin{small}
\begin{sc}
\begin{tabular}{lcccr}
\toprule
\textbf{Method} & \textbf{Gradient} & \textbf{Hessian} \\
\midrule
SGD    &$\mathbf{\hat{g}}_{t}$ \eqref{eq:g_est}, $\beta\equiv0$ & $\mathbf{I}$ & \\
SGD + Momentum & $\mathbf{\hat{g}}_{t}$
\eqref{eq:g_est} &  $\mathbf{I}$ \\
ADAM    & $\mathbf{\hat{g}}_{t}$ \eqref{eq:g_est} & $\hat{\mathbf{H}}_{t}$ \eqref{eq:Hessian_Adam}\\
ADAGRAD    & $\mathbf{\hat{g}}_{t}$ \eqref{eq:g_est}, $\beta\equiv0$ & $\hat{\mathbf{H}}_{t}$ \eqref{eq:Hessian_AdaGrad}\\
\bottomrule
\end{tabular}
\end{sc}
\end{small}
\end{center}
\vskip -0.1in
\end{table}

Our objective in this paper is to find the optimal values of $\alpha$ and $\beta$
at time $t$, which minimize the expected value of the loss function
$\hat{J}\left(\mathbf{w}\right)$ \eqref{eq:Taylor-3} when using
the update rule \eqref{eq:w_update-1}, i.e.,
\begin{equation}
\alpha_{Ot},\beta_{Ot}=\arg\underset{\alpha,\beta}{\min}\left(E\left\{ \hat{J}\left(\mathbf{w}_{t+1};\alpha,\beta\right)\right\} \right).\label{eq:oracle-1}
\end{equation}
We provide the solution for \eqref{eq:oracle-1} in the following
section. Recall that \eqref{eq:oracle-1} is solved for the general update rule \eqref{eq:w_update-1}. Thus, the proposed solution can be deployed in any SGD method that utilizes the gradient estimator $\mathbf{\hat{g}}_{t}$ \eqref{eq:g_est}.
As previously mentioned, the optimizers which appear in Table 1 are examined in the experiments section.

\section{Optimal Learning Rate and Momentum}\label{sec:optimal}

In this section, we derive the optimal learning rate and momentum as defined in \eqref{eq:oracle-1}. By changing variables such that $\alpha=1-\gamma_{1}$ and $\beta=\frac{\gamma_{2}}{1-\gamma_{1}}$,
we can rewrite the gradient estimator $\mathbf{\hat{g}}_{t}$ \eqref{eq:g_est}
as
\begin{equation}
\mathbf{\hat{g}}_{t}=\left(1-\gamma_{1}-\gamma_{2}\right)\mathbf{g}_{t}+\gamma_{2}\mathbf{\hat{g}}_{t-1}=\mathbf{g}_{t}-\mathbf{G}_{t}\boldsymbol{\gamma},\label{eq:g_est_1}
\end{equation}
where $\mathbf{G}_{t}$ is a $p\times2$ matrix defined as
\begin{equation}
\mathbf{G}_{t}=\left[\mathbf{g}_{t},\mathbf{g}_{t}-\mathbf{\hat{g}}_{t-1}\right],
\end{equation}
and $\boldsymbol{\gamma}=\left[\gamma_{1},\gamma_{2}\right]^{T}$ is a $2\times1$ vector.
Then, by substituting the update rule $\mathbf{w}_{t+1}$ \eqref{eq:w_update-1}
for $\mathbf{w}$ in \eqref{eq:Taylor-3} while using the gradient
estimator $\mathbf{\hat{g}}_{t}$ \eqref{eq:g_est_1}, we can rewrite
the expected value of the loss function \eqref{eq:Taylor-3} as
\begin{equation}
\begin{array}{c}
E\left\{ J\left(\mathbf{w}_{t+1};\boldsymbol{\gamma}\right)\right\} \approx E\left\{ \hat{J}\left(\mathbf{w}_{t+1};\boldsymbol{\gamma}\right)\right\} \\
=J\left(\mathbf{w}_{t}\right)-\mathbf{\bar{g}}_{t}^{T}E\left\{ \hat{\mathbf{H}}_{t}^{-1}\mathbf{g}_{t}\right\} +\frac{1}{2}E\left\{ \mathbf{g}_{t}^{T}\hat{\mathbf{H}}_{t}^{-1}\mathbf{\bar{H}}_{t}\hat{\mathbf{H}}_{t}^{-1}\mathbf{g}_{t}\right\} \\
-\boldsymbol{\gamma}^{T}\mathbf{b}_{t}+\frac{1}{2}\boldsymbol{\gamma}^{T}\mathbf{A}_{t}\boldsymbol{\gamma},
\end{array}\label{eq:Taylor-3-2-1-1}
\end{equation}
where
\begin{equation}
\mathbf{A}_{t}=E\left\{ \mathbf{G}_{t}^{T}\hat{\mathbf{H}}_{t}^{-1}\mathbf{\bar{H}}_{t}\hat{\mathbf{H}}_{t}^{-1}\mathbf{G}_{t}\right\} ,\label{eq:A_matrix}
\end{equation}
and
\begin{equation}
\mathbf{b}_{t}=E\left\{ \mathbf{G}_{t}^{T}\hat{\mathbf{H}}_{t}^{-1}\left(\left(\mathbf{g}_{t}-\mathbf{\bar{g}}_{t}\right)+\left(\mathbf{\bar{H}}_{t}\hat{\mathbf{H}}_{t}^{-1}-\mathbf{I}\right)\mathbf{g}_{t}\right)\right\} .\label{eq:b_vector}
\end{equation}
The matrix  $\mathbf{A}_{t}$ \eqref{eq:A_matrix} and the vector $\mathbf{b}_{t}$ \eqref{eq:b_vector}, are of size $2\times2$ and $2\times1$, respectively. The optimal vector $\boldsymbol{\gamma}$ at time $t$, which we denote
by $\boldsymbol{\gamma}_{Ot}=\left[\gamma_{1Ot},\gamma_{2Ot}\right]^{T}$,
is the solution that minimizes the loss function $E\left\{ \hat{J}\left(\mathbf{w}_{t+1};\boldsymbol{\gamma}\right)\right\} $
\eqref{eq:Taylor-3-2-1-1}, i.e.,
\begin{equation}
\boldsymbol{\gamma}_{Ot}=\arg\underset{\boldsymbol{\gamma}}{\min}\left(E\left\{ \hat{J}\left(\mathbf{w}_{t+1};\boldsymbol{\gamma}\right)\right\} \right)=\mathbf{A}_{t}^{-1}\mathbf{b}_{t}.\label{eq:oracle}
\end{equation}
Since $\boldsymbol{\gamma}_{Ot}$ \eqref{eq:oracle} depends on the
true gradient $\mathbf{\bar{g}}_{t}$ \eqref{eq:real_gradient}, which
is unknown in practice, we refer to it as the oracle solution. In
the following section we propose an estimator for the oracle solution
$\boldsymbol{\gamma}_{Ot}$ \eqref{eq:oracle}.

\section{Estimation of Oracle Solution}

\label{sec:estimation}

The oracle vector $\boldsymbol{\gamma}_{Ot}$ \eqref{eq:oracle} minimizes
the loss function $E\left\{ \hat{J}\left(\mathbf{w}_{t+1};\boldsymbol{\gamma}\right)\right\} $
\eqref{eq:Taylor-3-2-1-1}, but unfortunately depends on the unknown quantities  $\mathbf{A}_{t}$ \eqref{eq:A_matrix} and $\mathbf{b}_{t}$ \eqref{eq:b_vector}. Consider first the perplexing vector $\mathbf{b}_{t}$ \eqref{eq:b_vector} which depends on the unknown gradient $\mathbf{\bar{g}}_{t}$ \eqref{eq:real_gradient} and the Hessian matrix $\mathbf{\bar{H}}_{t}$
\eqref{eq:real_hessian}. With the aim of proceeding toward a practical method, we unfold the tangled equation by assuming that the Hessian is known, i.e.,  $\mathbf{\hat{H}}_{t}=\mathbf{\bar{H}}_{t}$, and provided by the optimizer currently in use (see Table 1). As a result, the vector $\mathbf{b}_{t}$ \eqref{eq:b_vector} can be simplified (after a few
mathematical manipulations) to
\begin{equation}
\mathbf{b}_{t}=\mathbf{1}_{2}V\left(\mathbf{g}_{t}|\mathbf{\hat{H}}_{t}\right),
\end{equation}
where
\begin{equation}
V\left(\mathbf{g}_{t}|\mathbf{\hat{H}}_{t}\right)=E\left\{ \left(\mathbf{g}_{t}-\mathbf{\bar{g}}_{t}\right)^{T}\mathbf{\hat{H}}_{t}^{-1}\left(\mathbf{g}_{t}-\mathbf{\bar{g}}_{t}\right)\right\} .\label{eq:Vgt}
\end{equation}
The scalar $V\left(\mathbf{g}_{t}|\mathbf{\hat{H}}_{t}\right)$ \eqref{eq:Vgt} still
depends on the unknown true gradient $\mathbf{\bar{g}}_{t}$\eqref{eq:real_gradient}, however, can be estimated. The derivation of an unbiased estimator of $V\left(\mathbf{g}_{t}|\mathbf{\hat{H}}_{t}\right)$
\eqref{eq:Vgt} appears in Appendix A, and is equal to
\begin{equation}
\hat{V}\left(\mathbf{g}_{t}|\mathbf{\hat{H}}_{t}\right)=\frac{\sum_{i=1}^{N}\left(\mathbf{g}_{t}^{\left(i\right)}-\mathbf{g}_{t}\right)^{T}\mathbf{\hat{H}}_{t}^{-1}\left(\mathbf{g}_{t}^{\left(i\right)}-\mathbf{g}_{t}\right)}{N\left(N-1\right)},\label{eq:Vgt_est}
\end{equation}
where $\mathbf{g}_{t}^{\left(i\right)}$ is the gradient produced
by the $i$-th observation within the current mini-batch of size $N$. The estimator of  $\mathbf{b}_{t}$ \eqref{eq:b_vector} is therefore 
\begin{equation}
\hat{\mathbf{b}}_{t}=\mathbf{1}_{2}\hat{V}\left(\mathbf{g}_{t}|\mathbf{\hat{H}}_{t}\right).\label{eq:b_vector_est}
\end{equation}
The estimator of  $\mathbf{A}_{t}$ \eqref{eq:A_matrix} is calculated by replacing
all expectations in $\mathbf{A}_{t}$ \eqref{eq:A_matrix} by their
sample counterparts, i.e., 
\begin{equation}
\hat{\mathbf{A}}_{t}= \mathbf{G}_{t}^{T}\hat{\mathbf{H}}_{t}^{-1}\mathbf{G}_{t}.\label{eq:A_matrix_est}
\end{equation}
Finally, by incorporating the estimators $\hat{\mathbf{b}}_{t}$ \eqref{eq:b_vector_est} and $\hat{\mathbf{A}}_{t}$ \eqref{eq:A_matrix_est}, the oracle solution $\boldsymbol{\gamma}_{Ot}$ \eqref{eq:oracle} is estimated by
\begin{equation}
\hat{\boldsymbol{\gamma}}_{Ot}=\hat{\mathbf{A}}_{t}^{-1}\hat{\mathbf{b}}_{t}.\label{eq:oracle_est}
\end{equation}
Consider that the values of $\hat{\boldsymbol{\gamma}}_{Ot}$ varies based on the current mini-batch at time $t$, we mitigates this effect by using an exponentially weighted moving average model  $\hat{\boldsymbol{\gamma}}_{Et}$, defined as 
\begin{equation}
\hat{\boldsymbol{\gamma}}_{Et}=\left(1-\upsilon\right)\hat{\boldsymbol{\gamma}}_{Ot} + \upsilon\hat{\boldsymbol{\gamma}}_{E(t-1)} .\label{eq:oracle_est_ewma}
\end{equation}
We summarize the proposed method for an automatic and simultaneous adjustment of the learning rate and momentum in Algorithm \ref{alg:AutoOpt}. The vector $\hat{\boldsymbol{\gamma}}_{Et}$ \eqref{eq:oracle_est_ewma} is computed in each step with a time-complexity that is equal or better than the time-complexity of the back-propagation algorithm \cite{rumelhart1986learning} (see Appendix B), therefore make the method feasible for a practical use in deep learning architectures. In the next section, we examine the proposed method for classification purposes using CNN's, and show that the technique attains the lowest or comparable classification errors as expected from theory. 
\begin{algorithm}
\textbf{Input:} 

1) Loss function $\bar{F}\left(\mathbf{w}\right)$ \eqref{eq:loss_F} with an initial weight vector
$\mathbf{w}_{0}$

2) Optimizer, based on the update rule (9). (e.g., Table 1).

\textbf{for} $t=1,\ldots,T$:
\begin{enumerate}
\item Calculate the oracle estimator $\hat{\boldsymbol{\gamma}}_{Ot}$ \eqref{eq:oracle_est}
\item Update $\hat{\boldsymbol{\gamma}}_{Et}$ \eqref{eq:oracle_est_ewma} using $\hat{\boldsymbol{\gamma}}_{Ot}$ \eqref{eq:oracle_est}
\item Calculate the gradient estimator $\mathbf{\hat{g}}_{t}$ \eqref{eq:g_est_1} with $\hat{\boldsymbol{\gamma}}_{Et}$ \eqref{eq:oracle_est_ewma}
\item Update the weight vector $\mathbf{w}_{t}$ according to $\mathbf{w}_{t+1}=\mathbf{w}_{t}-\hat{\mathbf{H}}_{t}^{-1}\mathbf{\hat{g}}_{t}$ \eqref{eq:w_update-1}, using the gradient estimator $\mathbf{\hat{g}}_{t}$ \eqref{eq:g_est_1}.
\end{enumerate}
\textbf{return} $\mathbf{w}_{T}$

\caption{AutoOpt: Automatic and Simultaneous Adjustment of Learning Rate and Momentum}
\label{alg:AutoOpt}
\end{algorithm}

\section{Experiments}
\label{sec:experiments}
\begin{figure}
  \centering
  \subfigure[]
  {\includegraphics[width=0.30\textwidth]{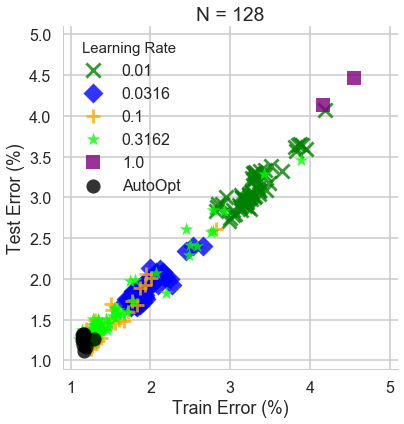}\label{fig:exph}}
  \subfigure[]
  {\includegraphics[width=0.30\textwidth]{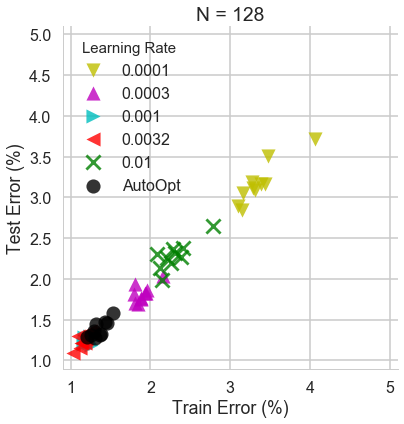}\label{fig:exgr}}
  \subfigure[]
  {\includegraphics[width=0.30\textwidth]{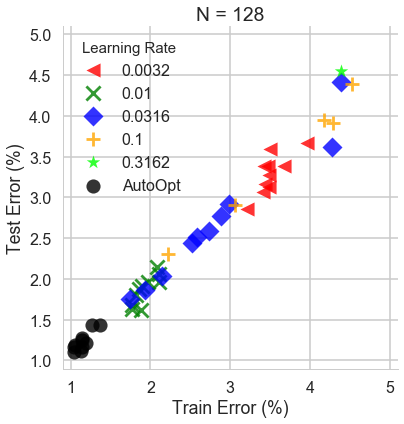}\label{fig:exgr}}
  
  \subfigure[]
  {\includegraphics[width=0.30\textwidth]{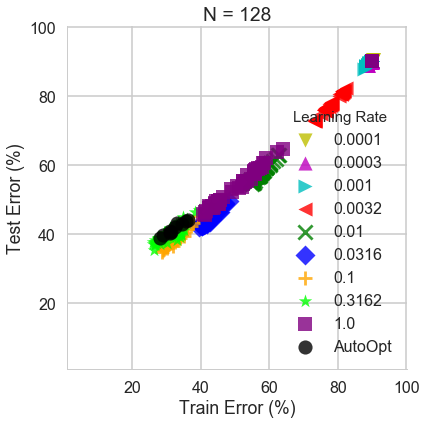}\label{fig:exph1}}
  \subfigure[]
  {\includegraphics[width=0.30\textwidth]{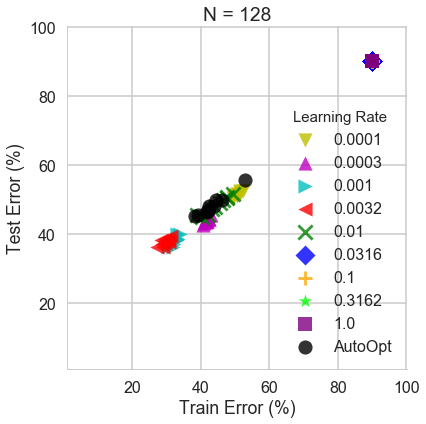}\label{fig:exgr1}}
  \subfigure[]
  {\includegraphics[width=0.30\textwidth]{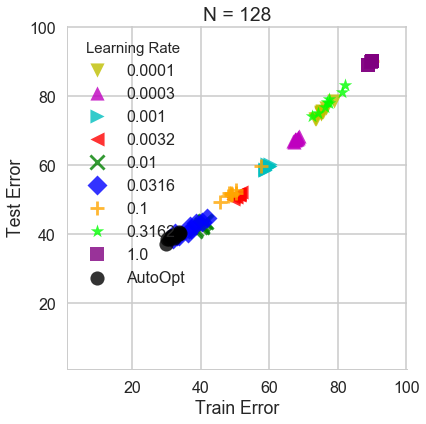}\label{fig:exgr2}}
  \caption{Train and test errors for $N=128$ after 10 epochs, for MNIST and CIFAR10, achieved by different optimizers. The MNIST classification errors, obtained by SGD, Adam, and AdaGrad, are presented in Figure (a), Figure (b), and Figure (c), respectively. The CIFAR10 errors, obtained by SGD, Adam, and AdaGrad, are presented in Figure (d), Figure (e), and Figure (f), respectively. The proposed method (denoted in black dots) automatically achieves comparable, or the lowest errors, and therefore avoids the exhaustive manual tuning.}
  \label{fig:TrainTestCIFAR128}
\end{figure}

We utilize the proposed method to train CNN classifiers for the MNIST \cite{mnistlecun} and CIFAR10 \cite{krizhevsky2009learning} data sets.
%The method implementation and experiments are provided in \cite{AutoOpt}.
The neural network architecture for MNIST has two convolution layers (10 and 20 channels with a kernel size 5), max pooling (kernel size 2), ReLU non-linearity and drop-out, which produces 320 features, followed by two fully connected layers (50 and 10 output features). The output layer is a log softmax layer, and the loss function is the \emph{negative log likelihood} 
(NLL) loss.
The architecture for CIFAR10 (3 input channels) has two convolution layers (6 and 16 channels with a kernel size 5), max pooling (kernel size 2) and ReLU non-linearity, which produces 400 features followed by three fully connected layers (120, 84, and 10 output features). The output layer is again a log softmax layer and the loss function is the NLL loss. We run thousands of configurations with different learning rate, momentum and random initial weights that were generated from 10 different seeds.
For the SGD optimizer we run exhaustive hyperparameter search with {\em learning rate} $\in L = \{10^{-4}, \allowbreak 10^{-3.5}, 10^{-3}, 10^{-2.5}, 10^{-2}, 10^{-1.5}, 10^{-1}, 10^{-0.5}, 1\}$, 
{\em momentum} $\in M = \{0, 0.3, 0.8, 0.9, 0.95, 0.99\}$ and
{\em batch size} $\in N = \{8, 16, 32, 64, 128, \allowbreak 256\}$ values.
This yields $|L| * |M| * |N| = 324$ different settings for each data set.
Every setting is run with 10 different seeds.
Thus, a total of 3,240 different configurations have been tested for the SGD optimizer. For the Adam and AdaGrad optimizers, best parameter search is carried out among the same 
{\em learning rate} and {\em batch size} values as in SGD. In Adam, we use $\beta_1=0.9$ and $\beta_2=0.99$ as suggested in the original Adam paper. The parameter tuning for Adam and AdaGrad optimizers each yield a total of 54 different settings and 540 results due to the use of 10 different seeds. 
\begin{figure}
  \centering
  \subfigure[]
  {\includegraphics[width=0.49\textwidth]{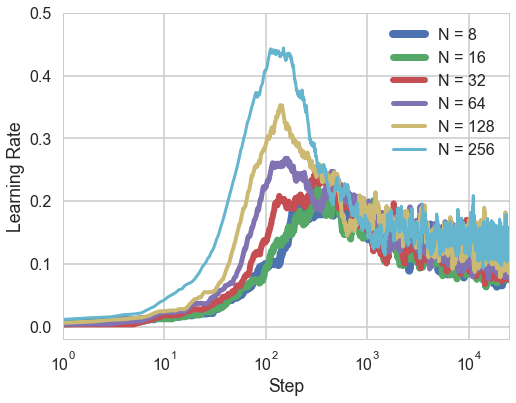}\label{fig:exph}}
  \subfigure[]
  {\includegraphics[width=0.49\textwidth]{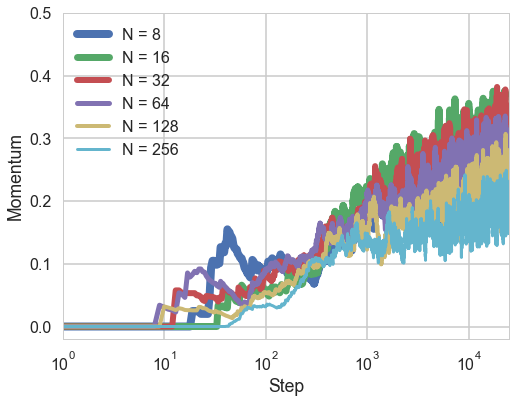}\label{fig:exgr}}
  \caption{Automatic learning rate and momentum for SGD as a function of step for the first convolutional layer of the MNIST classifier. As expected, the learning rate increases as the mini-batch size increases. Whereas the learning rate starts to decay automatically, the previous gradient shifts to be less biased. As a result, the method utilizes the benefits of the previous gradient by gradually increasing the momentum value.}
  \label{fig:LearningRate}
\end{figure}
The proposed method produces for each layer its own learning rate and momentum. To get further insights and intuition regarding the approach's behavior; we present in Figure \ref{fig:LearningRate} the learning rate and momentum, generated by SGD, for the first convolutional layer of the MNIST classifier as a function of the mini-batch size $N$. As expected, the learning rate, provided in Figure \ref{fig:LearningRate}(a), increases as the mini-batch size increases. Since a larger mini-batch result with less variance of the gradient estimator, the proposed method increases the learning rate. Once the learning rate reaches its peak, we can observe a learning rate decay. The learning rate decay is another hyperparameter that should be set in advance. In our case, however, the learning rate decay emerges automatically based on the data. The momentum values are provided in Figure \ref{fig:LearningRate}(b). Initially, since the learning rate is relatively large, the previous gradient $\mathbf{\hat{g}}_{t-1}$ is too biased to be combined with the current gradient $\mathbf{\hat{g}}_{t}$, which results with low values of momentum. As the learning rate decay (and step size become smaller), the previous gradient $\mathbf{\hat{g}}_{t-1}$ is less biased and therefore gain a greater presence when combined with the current unbiased gradient estimator $\mathbf{g}_{t}$ \eqref{eq:gt}. The latter is reflected by increasing values of momentum as can be seen in Figure \ref{fig:LearningRate}(b).
\begin{figure}
  \centering
  \subfigure[]
  {\includegraphics[width=0.32\textwidth]{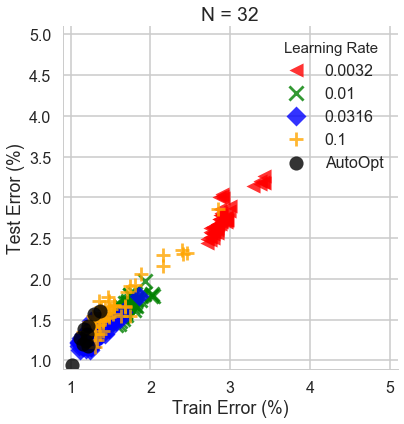}\label{fig:exph}}
  \subfigure[]
  {\includegraphics[width=0.32\textwidth]{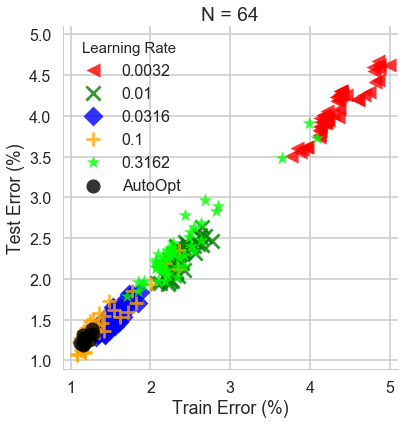}\label{fig:exgr}}
  \subfigure[]
  {\includegraphics[width=0.32\textwidth]{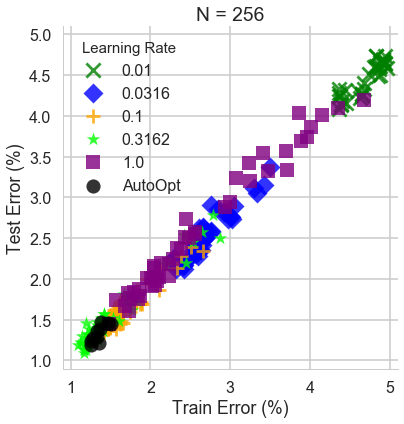}\label{fig:exgr}}
  \caption{Train and test errors for the MNIST CNN classifier, after 10 epochs with SGD, as a function of the mini-batch size. The proposed method (denoted in black dots) achieves comparable, or the lowest error, which otherwise would be attained by an exhaustive manual tuning.}
  \label{fig:TrainTestMNIST_SGD}
\end{figure}
\begin{table}[]
    \scriptsize 
    \centering
    \caption{Lowest train and test errors (mean and std across 10 seeds), for SGD, Adam, and AdaGrad, obtained by a tedious manual tuning. The optimal learning rate and momentum are provided by the first and second configuration values, respectively. The automatic tuning counterparts of these optimizers achieve comparable, or better results.}
    \label{tab:min_test_error_v3}
\begin{tabular}{rrrrrrrr}
\toprule
 & & \multicolumn{1}{c}{\tiny\textbf{SGD}} &  \multicolumn{1}{c}{\tiny\textbf{AutoSGD}} & \multicolumn{1}{c}{\tiny\textbf{Adam}} & \multicolumn{1}{c}{\tiny\textbf{AutoAdam}} & \multicolumn{1}{c}{\tiny\textbf{AdaGrad}} & \multicolumn{1}{c}{\tiny\textbf{AutoAdaGrad}} \\
\midrule[0.2pt]
\multicolumn{2}{c}{\textbf{\textit{\# of configurations}}} & \multicolumn{1}{c}{\tiny\textbf{324}} & \multicolumn{1}{c}{\tiny\textbf{1}} & \multicolumn{1}{c}{\tiny\textbf{54}} & \multicolumn{1}{c}{\tiny\textbf{1}} & \multicolumn{1}{c}{\tiny\textbf{54}} & \multicolumn{1}{c}{\tiny\textbf{1}} \\
\midrule
\multirow{3}{*}{\textbf{MNIST}}   &  \emph{train}   &    1.17\%, 0.14\% & 1.10\%, 0.05\% &   1.05\%, 0.05\% & 1.34\%, 0.11\% & 1.75\%, 0.38\% &  1.15\%, 0.10\% \\
                                  &  \emph{test}    &    1.21\%, 0.11\% & 1.22\%, 0.16\% &   1.14\%, 0.04\% & 1.35\%, 0.12\% & 1.86\%, 0.18\% &  1.23\%, 0.12\% \\
                                  &  \emph{config.} &    $10^{-2}$, 0.0 &        AutoOpt & $10^{-3.5}$, 0.9 &        AutoOpt & $10^{-2}$      &         AutoOpt \\
\midrule[0.2pt]
\multirow{3}{*}{\textbf{CIFAR10}} &  \emph{train}   &    26.4\%, 1.39\% &  30.0\%, 1.33\% &   27.2\%, 1.58\% &  42.6\%, 3.29\% & 31.4\%, 2.92\% & 27.3\%, 1.44\% \\
                                  &  \emph{test}    &    36.4\%, 0.85\% &  38.7\%, 1.56\% &   36.4\%, 1.03\% &  47.8\%, 2.69\% & 38.3\%, 1.31\% & 36.7\%, 0.82\% \\
                                  &  \emph{config.} & $10^{-1.5}$, 0.99 &         AutoOpt & $10^{-3.5}$, 0.9 &         AutoOpt & $10^{-1.5}$    &        AutoOpt \\
\bottomrule
\end{tabular}
\end{table}
In Figure \ref{fig:TrainTestMNIST_SGD}, we present the scatter plots of test errors vs. train errors, achieved by the CNN architecture for MNIST after 10 epochs. It can be seen how the optimal learning rate changes as a function of the mini-batch size while the method adapts and achieves the lowest error rates. For $N=32$, we observe that the optimal learning rate is $\alpha=0.031$, denoted by blue diamonds. It can be seen that the proposed method (denoted in the figure as AutoOpt), achieves the same error rates by automatically adapt to the optimal learning rate. For $N=64$, the learning rate $\alpha=0.031$ becomes sub-optimal and the new optimal learning rate increases to $\alpha=0.1$. Still, the proposed method adapts and achieves comparable test and train errors. Finally, when the mini-batch size increases to $N=256$, the optimal learning rate increases to $\alpha=0.31$, and the previous learning rate of $\alpha=0.1$ become sub-optimal. The optimal learning rate for $N=32$ ($\alpha=0.031$) is almost off the plot for the case of $N=256$. Likewise, the proposed method performs similarly, or better, in the case of Adam and AdaGrad, as observed in Figure \ref{fig:TrainTestCIFAR128}. For completeness, the lowest train and test errors (mean and std from 10 different seeds) obtained by the best performing configuration are summarized in Table~\ref{tab:min_test_error_v3}. The first row in Table~\ref{tab:min_test_error_v3} lists the total number of parameter settings per data set for each optimizer. It can be observed that by conducting a tedious manual tuning, SGD, Adam and AdaGrad achieve comparable results. The deployment of the proposed method for these optimizers (AutoSGD, AutoAdam and AutoAdaGrad), avoids the tedious manual tuning while still achieving comparable train and test errors as shown in the table. The case of AutoAdam with CIFAR10 have a relatively large gap of more than 10\%. The reason for that gap is a discrepancy between the unknown Hessian and the Hessian estimator provided by Adam (See the method's assumption regarding the Hessian in Section 4, and Table 1 for different Hessian estimators). Nevertheless, as observed in Figure \ref{fig:TrainTestCIFAR128}(e), the automatically tuned version attains a near-optimal solution in comparison to other configurations that were examined, and therefore, saves a substantial amount of time and efforts. Moreover, since SGD, Adam, and AdaGrad perform relatively the same after carefully tuned, it is straightforward to switch between their automatically tuned counterparts and reveal the best performing optimizer. An examination of three different automatic optimizers is still easily managed, in comparison to an intensive manual adjustment of each optimizer alone. The difference between the number of settings for the original (manually tuned) optimizers and their automatic counterparts show the huge gain of the approach. To conclude, in all cases the proposed method automatically attains the lowest, or comparable test and train errors that otherwise would be achieved by an exhaustive manual tuning. 

\section{Conclusions} 
\label{sec:conclusions}
We tackle the manual tuning problem of two imperative hyperparameters: the learning rate and momentum. In Section \ref{sec:optimal}, we derive a general method to compute the optimal learning rate and momentum
that minimize the expected loss \eqref{eq:oracle-1} after the next
update. The technique relies on the unbiased gradient estimator $\mathbf{g}_{t}$
\eqref{eq:gt} which depends on the current mini-batch of size $N$
at time $t$, and is summarized in Algorithm 1. The method is generic and can easily be deployed for different optimizers. Specifically, we examined the method for three well-known optimizers: SGD, Adam, and AdaGrad. The experimental results in Section 5 confirm the theoretical expectations where the learning rate and momentum automatically tuned to maximally decrease the expected loss by utilizing the mini-batch statistics, thus eliminating the need for an exhaustive manual tuning. We show that the method either outperform or comparable to the manual tuning by comparing classification errors of CNN based classifiers.  Given the successful validation and the method's generality, we intend to expand the proposed method into other deep learning architectures, and deploy it into more state-of-the-art optimizers previously mentioned. Also, as the optimal values of learning rate and momentum can freely increase or decrease based on the available data, we intend
to examine the proposed approach for on-line training scenarios with
non-stationary data. In these cases, the learning rate and
momentum automatically adapt to the evolving data, and may stabilize to more appropriate settings based on the new distribution. 

\bibliography{autoopt}
\bibliographystyle{unsrt}

\section*{Appendix A: Derivation of $\hat{V}\left(\mathbf{g}_{t}|\mathbf{\hat{H}}_{t}\right)$
\eqref{eq:Vgt_est}}

The expression $V\left(\mathbf{g}_{t}|\mathbf{\hat{H}}_{t}\right)$
\eqref{eq:Vgt} can be written as
\[
\begin{array}{c}
V\left(\mathbf{g}_{t}|\mathbf{\hat{H}}_{t}\right)=E\left\{ \left(\mathbf{g}_{t}-\mathbf{\bar{g}}_{t}\right)^{T}\mathbf{\hat{H}}_{t}^{-1}\left(\mathbf{g}_{t}-\mathbf{\bar{g}}_{t}\right)\right\} \\
=E\left\{ \left(\frac{1}{N}\sum_{i=1}^{N}\mathbf{g}_{t}^{\left(i\right)}-\mathbf{\bar{g}}_{t}\right)^{T}\mathbf{\hat{H}}_{t}^{-1}\left(\frac{1}{N}\sum_{i=1}^{N}\mathbf{g}_{t}^{\left(i\right)}-\mathbf{\bar{g}}_{t}\right)\right\} \\
=\frac{1}{N^{2}}\sum_{i=1}^{N}E\left\{ \left(\mathbf{g}_{t}^{\left(i\right)}-\mathbf{\bar{g}}_{t}\right)^{T}\mathbf{\hat{H}}_{t}^{-1}\left(\mathbf{g}_{t}^{\left(i\right)}-\mathbf{\bar{g}}_{t}\right)\right\} \\
=\frac{\sum_{i=1}^{N}E\left\{ \left(\mathbf{g}_{t}^{\left(i\right)}-\mathbf{g}_{t}\right)^{T}\mathbf{\hat{H}}_{t}^{-1}\left(\mathbf{g}_{t}^{\left(i\right)}-\mathbf{g}_{t}\right)\right\} }{N^{2}} + \frac{1}{N} V\left(\mathbf{g}_{t}|\mathbf{\hat{H}}_{t}\right),
\end{array}
\]
and finally
\[
\begin{array}{c}
V\left(\mathbf{g}_{t}|\mathbf{\hat{H}}_{t}\right)\left(1-\frac{1}{N}\right)=\frac{\sum_{i=1}^{N}E\left\{ \left(\mathbf{g}_{t}^{\left(i\right)}-\mathbf{g}_{t}\right)^{T}\mathbf{\hat{H}}_{t}^{-1}\left(\mathbf{g}_{t}^{\left(i\right)}-\mathbf{g}_{t}\right)\right\} }{N^{2}}\\
V\left(\mathbf{g}_{t}|\mathbf{\hat{H}}_{t}\right)=E\left\{\frac{\sum_{i=1}^{N} \left(\mathbf{g}_{t}^{\left(i\right)}-\mathbf{g}_{t}\right)^{T}\mathbf{\hat{H}}_{t}^{-1}\left(\mathbf{g}_{t}^{\left(i\right)}-\mathbf{g}_{t}\right) }{N\left(N-1\right)}\right\}
\end{array}
\]
The expression in the expectation is therefore the unbiased estimator of $V\left(\mathbf{g}_{t}|\mathbf{\hat{H}}_{t}\right)$
\eqref{eq:Vgt}, provided in $\hat{V}\left(\mathbf{g}_{t}|\mathbf{\hat{H}}_{t}\right)$
\eqref{eq:Vgt_est}. When the Hessian estimator $\mathbf{\hat{H}}_{t}$ is diagonal, as in SGD, Adam, and AdaGrad (see Table 1), the estimator $\hat{V}\left(\mathbf{g}_{t}|\mathbf{\hat{H}}_{t}\right)$
\eqref{eq:Vgt_est} have a time-complexity of $O\left(pN\right)$.

\section*{Appendix B: Algorithm 1 -   Time Complexity for Deep Learning Architectures}
The proposed method for an automatic adjustment of learning rate and momentum (see Algorithm 1) can be utilized in \emph{fully-connected neural networks} (FCNN's) and \emph{convolution neural networks} (CNN's). Practically, when the Hessian estimator is a diagonal matrix, as in SGD, Adam, and AdaGrad (see Table 1) , the time complexity of the proposed algorithm remains the same as for the back-propagation phase \cite{rumelhart1986learning}, thus doesn't increase the time complexity of the training. 
Consider the case of a
FCNN with $L$ layers \cite{goodfellow2016deep}.
The $L$ layers, $l=1,\ldots,L$, are connected to each other by the
following relation:
\begin{equation}
\mathbf{A}^{[l]}=f(\mathbf{Z}^{[l]})=f(\mathbf{W}^{[l]}\mathbf{A}^{[l-1]}+\mathbf{b}^{[l]}\mathbf{e}^{T})
\end{equation}
where $f(.)$ is the activation function (for example, $sigmoid()$
or \emph{rectified linear unit} $ReLU(.)$\cite{nair2010rectified}).
The weight matrices $\mathbf{W}^{[l]}$ and the bias vectors $\mathbf{b}^{[l]}$,
are of size $p^{\left[l\right]}\times p^{\left[l-1\right]}$, and
$p^{\left[l\right]}\times1$, respectively. The activation matrix
$\mathbf{A}^{[l]}$ (as well as $\mathbf{Z}^{[l]}$) is of size $p^{\left[l\right]}\times N$,
where $N$ is the mini-batch size. Note that $\mathbf{A}^{[0]}$ is
the forwarded mini-batch of size $p^{\left[0\right]}\times N$.

The gradients of a loss function $\mathcal{\mathcal{J}}$ with respect
to the parameters $\left\{ \mathbf{W}^{[l]},\mathbf{b}^{[l]}\right\} ,l=1,\ldots,L$
are calculated using the back-propagation algorithm \cite{rumelhart1986learning},
i.e., we can compute the gradients $\left\{ \mathbf{G}^{[l]},d\mathbf{b}^{[l]}\right\} $,
$l=1,\ldots,L$ by the following procedure:

For $l=L,\ldots,1$ calculate:
\begin{equation}
d\mathbf{Z}^{[l]}=\frac{\partial\mathcal{\mathcal{J}}}{\partial\mathbf{Z}^{[l]}}=d\mathbf{A}^{[l]}\circ f'(\mathbf{Z}^{[l]}).\label{eq:dz}
\end{equation}
\begin{equation}
\mathbf{G}^{[l]}=\frac{\partial\mathcal{\mathcal{J}}}{\partial\mathbf{W}^{[l]}}=\frac{1}{N}d\mathbf{Z}^{[l]}\mathbf{A}^{[l-1]T},\label{eq:dw_backprop}
\end{equation}
\begin{equation}
d\mathbf{b}^{[l]}=\frac{\partial\mathcal{\mathcal{J}}}{\partial\mathbf{b}^{[l]}}=\frac{1}{N}\sum_{i=1}^{N}d\mathbf{Z}^{[l](i)},\label{eq:db}
\end{equation}
\begin{equation}
d\mathbf{A}^{[l-1]}=\frac{\partial\mathcal{\mathcal{J}}}{\partial\mathbf{A}^{[l-1]}}=\mathbf{W}^{[l]T}d\mathbf{Z}^{[l]}\label{eq:dA_prev}.
\end{equation}
The time complexities for \eqref{eq:dz}, \eqref{eq:dw_backprop}, \eqref{eq:db}, \eqref{eq:dA_prev} are  $O\left(p^{\left[l\right]}N\right)$, $O\left(p^{\left[l\right]}p^{\left[l-1\right]}N\right)$, $O\left(p^{\left[l\right]}N\right)$, and $O\left(p^{\left[l\right]}p^{\left[l-1\right]}N\right)$, respectively. 
Recall the computation of the gradient $\mathbf{G}^{[l]}$ \eqref{eq:dw_backprop}, the individual gradients $\mathbf{G}^{[l](i)},i=1,2,...,N$ are calculated by 
\begin{equation}
\mathbf{G}^{[l](i)}=\mathbf{z}_{i}^{[l]}\mathbf{a}_{i}^{[l-1]T},\label{eq:dw_backprop_individuals}
\end{equation}
where $\mathbf{G}^{[l](i)}$ have a time complexity of $O\left(p^{\left[l\right]}p^{\left[l-1\right]}\right)$. Therefore, the time complexity for all individual gradients is $O\left(p^{\left[l\right]}p^{\left[l-1\right]}N\right)$.
The time complexity for $\hat{V}\left(\mathbf{G}^{[l]}|\mathbf{\hat{H}}_{t}\right)$ \eqref{eq:Vgt_est} when the Hessian estimator $\mathbf{\hat{H}}_{t}$ is diagonal is also the same as computing $\mathbf{G}^{[l]}$
\eqref{eq:dw_backprop}, i.e., $O\left(p^{\left[l\right]}p^{\left[l-1\right]}N\right)$ (see Appendix A). Therefore, the time complexity of Algorithm 1, when deployed for a fully-connected layer, have the same time complexity as its back-propagation, which is $O\left(p^{\left[l\right]}p^{\left[l-1\right]}N\right)$ for the $l$-th layer.

Similarly, assume that the $l$-th layer is a convolutional layer having the filter $\mathbf{W}^{\left[l\right]}$. The latter is a 4-dimensional tensor of size $f\times f\times C^{\left[l-1\right]}\times C^{\left[l\right]}$
where $f\times f$ is the convolve window size, and $C^{\left[l\right]}$
denotes the number of channels of the $l$-th layer.  The gradient
of the $c$-th channel ($c=1,\ldots,C^{\left[l\right]}$) in $\mathbf{W}^{\left[l\right]}$ is a 3-dimensional tensor of size $f\times f\times C^{\left[l-1\right]}$, denoted as $d\mathbf{W}_{c}^{\left[l\right]}$, and computed
by 
\begin{equation}
d\mathbf{W}_{c}^{\left[l\right]}=\frac{1}{N}\sum_{i=1}^{N}d\mathbf{W}_{c}^{\left[l\right]\left(i\right)},\label{eq:dW_conv-1}
\end{equation}
where $d\mathbf{W}_{c}^{\left[l\right]\left(i\right)}$ is the gradient with respect to the $i$-th observation, defined as
\begin{equation}
d\mathbf{W}_{c}^{\left[l\right]\left(i\right)}=\sum_{h=0}^{H^{\left[l\right]}}\sum_{w=0}^{W^{\left[l\right]}}\mathbf{A}_{hw}^{\left[l-1\right]\left(i\right)}dz_{hw}^{\left[l\right]\left(i\right)}.\label{eq:dW_conv_ind}
\end{equation}
The scalar $dz_{hw}^{\left[l\right]\left(i\right)}$ indicates the gradient of the activation $z_{hw}^{\left[l\right]\left(i\right)}$, and $H^{\left[l\right]},W^{\left[l\right]}$ denote the height and
width of the $l$-th layer.
The tensor $\mathbf{A}_{hw}^{\left[l-1\right]\left(i\right)}$ of size $f\times f\times C^{\left[l-1\right]}$ corresponds to the activation of the previous layer which was used to generate the activation $z_{hw}^{\left[l\right]\left(i\right)}$ of the $i$-th observation.

Therefore, the time-complexity for calculating the mini-batch
gradient $d\mathbf{W}_{c}^{\left[l\right]}$ \eqref{eq:dW_conv-1} (or the set of individual gradients $\left\{ d\mathbf{W}_{c}^{\left[l\right]\left(i\right)}\right\} _{i=1}^{N}$
\eqref{eq:dW_conv_ind}) is $O\left(f^{2}C^{\left[l-1\right]} C^{\left[l\right]}NH^{\left[l\right]}W^{\left[l-1\right]}\right)$. Once we have the gradients per-observation, we can proceed to Algorithm 1 that have a time-complexity of $O\left(f^{2}C^{\left[l-1\right]} C^{\left[l\right]}N\right)$, which is lower than the time-complexity of $d\mathbf{W}_{c}^{\left[l\right]}$ \eqref{eq:dW_conv-1}.

\end{document}